%% file: acl_latex.tex
\pdfoutput=1

\documentclass[11pt]{article}

\usepackage[]{acl}

\usepackage{times}
\usepackage{latexsym}

\usepackage{hyperref}
\usepackage{graphicx}
\usepackage{enumitem}
\usepackage{booktabs}
\usepackage{multirow}
\usepackage{comment}
\usepackage{cleveref}
\usepackage{tabularx}
\usepackage{fontawesome5}
\crefformat{footnote}{#2\footnotemark[#1]#3}
\usepackage{pdfpages}

\usepackage[LAE,T1]{fontenc}
\usepackage[utf8]{inputenc}
\usepackage{arabtex}

\renewenvironment{abstract}%
		 {\centerline{\large\bf Abstract}%
		  \begin{list}{}%
		     {\setlength{\rightmargin}{0.6cm}%
		      \setlength{\leftmargin}{0.6cm}}%
		   \item[]\ignorespaces%

            \fontsize{10pt}{12pt}\selectfont
		   }%
		 {\unskip\end{list}}

\usepackage{utf8}
\setcode{utf8}

\usepackage{microtype}

\usepackage[colorinlistoftodos,prependcaption,textsize=tiny]{todonotes}

\newcommand{\tcc}{\textsc{Tcc}}
\newcommand{\Sim}{\textsc{Sim}}

\title{Eliciting Better Multilingual Structured Reasoning\\ from LLMs through Code}

\author{Bryan Li$^1$\Thanks{ Work done during an internship at Amazon}
, Tamer Alkhouli$^2$\Thanks{ corresponding author}, Daniele Bonadiman$^2$, Nikolaos Pappas$^2$, Saab Mansour$^2$ \\
  $^1$University of Pennsylvania, bryanli@seas.upenn.edu\\ $^2${\raisebox{-0.5ex}{\makebox[17pt][l]{\Large \faAws }} AI Labs}, \{alkhouli, dbonadim, nppappa, saabm\}@amazon.com}

\begin{document}
\maketitle
\begin{abstract}
The development of large language models (LLM) has shown progress on reasoning, though studies have largely considered either English or simple reasoning tasks. To address this, we introduce a multilingual structured reasoning and explanation dataset, termed xSTREET, that covers four tasks across six languages. xSTREET exposes a gap in base LLM performance between English and non-English reasoning tasks.\footnote{\url{https://github.com/amazon-science/xstreet} released under CC-BY-4.0.}

We then propose two methods to remedy this gap, building on the insight that LLMs trained on code are better reasoners. First, at training time, we augment a code dataset with multilingual comments using machine translation  while keeping program code as-is. Second, at inference time, we bridge the gap between training and inference by employing a prompt structure that incorporates step-by-step code primitives to derive new facts and find a solution. 
Our methods show improved multilingual performance on xSTREET, most notably on the scientific commonsense reasoning subtask. Furthermore, the models show no regression on non-reasoning tasks, thus demonstrating our techniques maintain general-purpose abilities.
\end{abstract}

\section{Introduction}

\begin{figure}[t!]
    \centering
    \includegraphics[width=.95\linewidth]{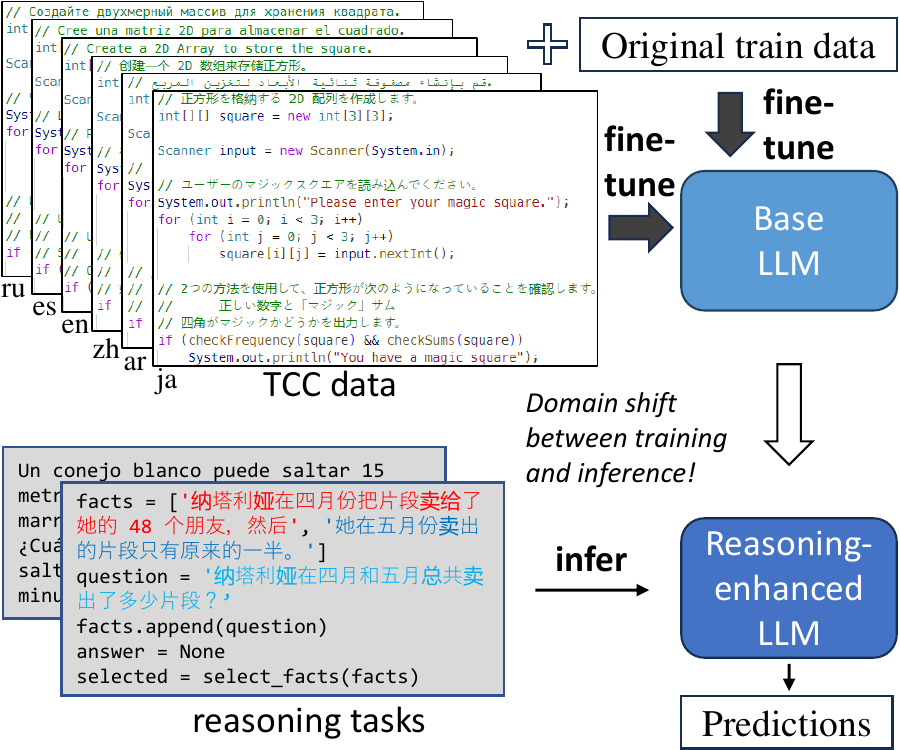}
    \caption{An overview of our methods to improve multilingual structured reasoning. First (top), we create the translated code comments (\tcc) dataset, and use it in a fine-tuning setup. Second (bottom), we use the resulting LLM for inference on reasoning tasks. We find the most success with a code prompt format that bridges the representations between training and inference.}
    \label{fig:recipe}
\end{figure}

\begin{figure*}[ht!]
    \centering
    \includegraphics[width=0.8\linewidth]{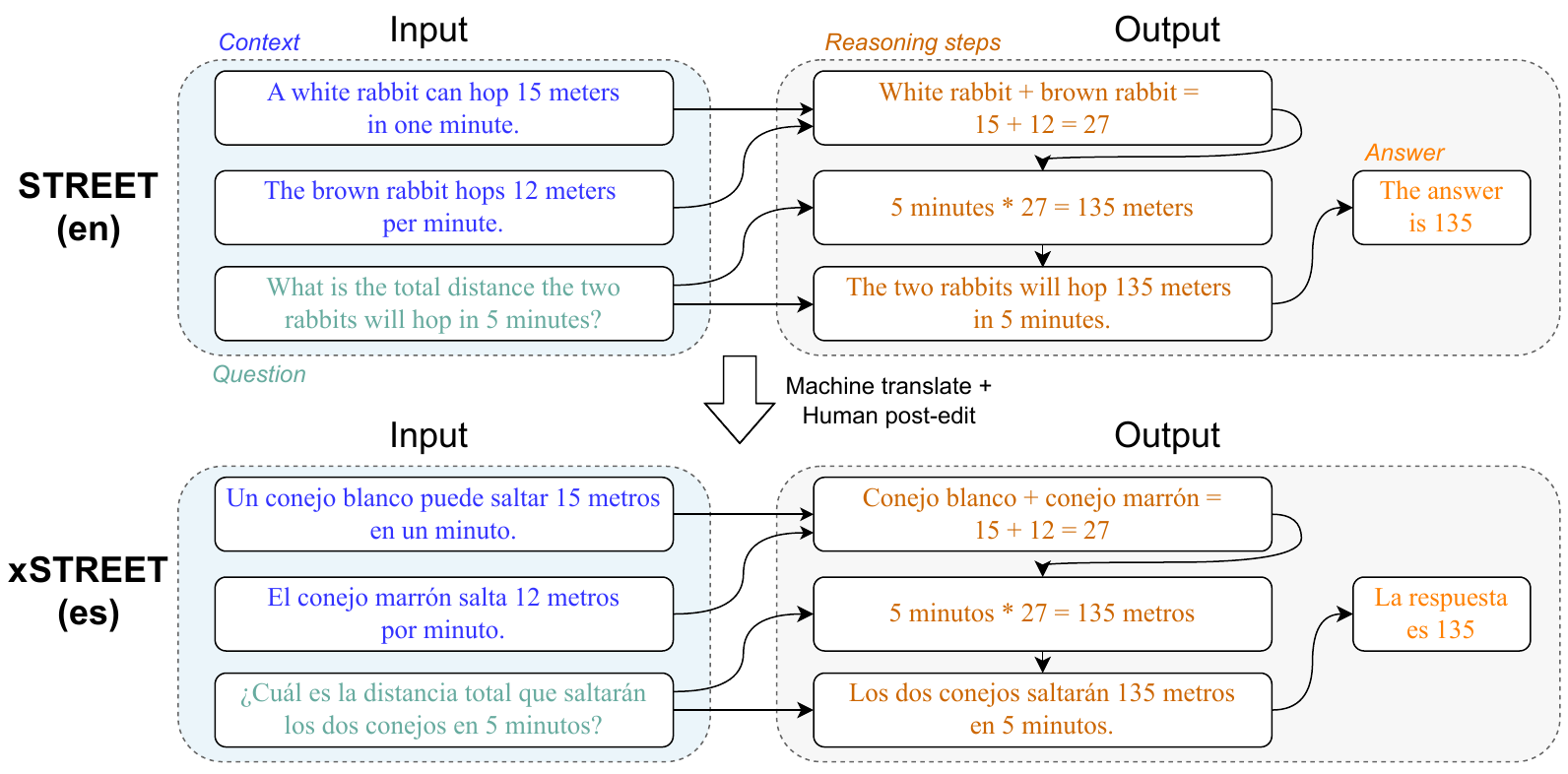}
    \caption{The translation process for an xSTREET entry. We start from an example from STREET~\cite{ribeiro2022street}. The reasoning graphs are directly transferred, while each sentence text is translated. Note that this shows only one (of 4) task, GSM8K, and one (of 5) language, Spanish.}
    \label{fig:street_task}
\end{figure*}

The ability to perform complex reasoning tasks is fundamental to human intelligence, where multiple steps of thought are required. Complex reasoning remains an open-problem for large language models (LLMs), despite some recent progress. Prior works consider complex reasoning tasks specified only in English. Such an English-centric perspective provides a limited assessment of the underlying reasoning capabilities of LLMs, given any specific language is largely a surface-form representation. This motivates our first inquiry into the multilingual complex reasoning capabilities of LLMs.

We introduce the xSTREET reasoning and explanation dataset (as shown in Figure~\ref{fig:street_task}). xSTREET covers 4 tasks, and extends the English STREET benchmark~\cite{ribeiro2022street} to 5 additional diverse languages, inheriting the source's expert annotations and structured graphs for reasoning steps (7.8 average steps/answer). The tasks cover arithmetic, logic and science commonsense problems. We perform machine translation for the training and development data splits, and also perform human post-editing to the test sets, to ensure a high quality multilingual benchmark. We use xSTREET to evaluate several LLMs, identifying the multilingual setting as significantly challenging.

To remedy the non-English reasoning gap, we turn to the widely accepted hypothesis that LLMs trained on code are better at reasoning than those trained only on text. This \textit{code and reasoning hypothesis} has been empirically corroborated by several papers~\cite{Suzgun2022ChallengingBT,Liang2023HolisticEO, Hendy2023HowGA}. Our work takes a further step in investigating the extent to which this hypothesis holds for non-English tasks. We proceed with the insight that code can be leveraged as a structured framework to represent the underlying reasoning steps, regardless of the surface-form language of the task. 
We thus propose two techniques to elicit better multilingual complex reasoning from LLMs (as shown in Figure~\ref{fig:recipe}): at training time through a lightweight fine-tuning recipe on code, and at inference time using a novel code prompt format.

In the LLM literature, many capabilities have been characterized as `emergent' with model scale~\cite{wei2022emergent,patel2022bidirectional}. Recent work on complex reasoning has thus focused on huge (175B+) and closed-source models. In our work, we instead aim to boost performance on far smaller open-source LLMs (7B). To make our findings reproducible, we release our benchmark.
Our contributions are:

\begin{enumerate}[noitemsep,topsep=0pt]
    \item We collect and release the first \textbf{dataset for multilingual structured reasoning, xSTREET}, covering 6 diverse languages and 4 tasks (5.5K entries total).
    \item At train time: we enhance reasoning capabilities of off-the-shelf LLMs by further training on \textbf{program code data where code is interleaved with non-English comments}. To this end, we augment a source code corpus through translating code comments and apply low-rank parameter-efficient fine-tuning (LoRA ~\cite{hu2021lora}) to BLOOMZ~\cite{muennighoff2022crosslingual}. Our method is effective yet lightweight, while preserving general-purpose LM capabilities.
 
    \item At inference time: we design a code-like prompting format that mimics the structure of the reasoning tasks by \textbf{interweaving function calls and multilingual text}. We show this format outperforms several other prompt formats used.
    \item We evaluate multiple LLMs (BLOOMZ, GPT-3, Falcon-40b-instruct) on our benchmark, and show improved performance — even for top-performing models — across structured reasoning tasks in different languages. As our inference and training-time techniques are orthogonal, we show that they can be used in tandem to achieve the best performance.    
    \item We perform qualitative and quantitative analysis to understand the roles of both the program code, and the code comments. Our findings taken together suggest that code elicits better multilingual structured reasoning by improving LLM's adherence to the reasoning format.
    
\end{enumerate}

\section{Related Work}

We adopt the scope of complex reasoning from~\citet{ribeiro2022street}, and use complex reasoning and structured reasoning interchangeably. The goal is to study the reasoning process itself, and how an LLM can structure its output in steps to improve the final performance. The tasks are selected so that the knowledge to answer a question is contained with the input question and context. For ease of evaluation, the answers are multiple-choice selections, or numbers for arithmetic reasoning.
Further details are given in \S\ref{sec:street}. 

\subsection{Code \& Reasoning Hypothesis for LLMs}
This hypothesis arose from empirical evidence by several concurrent works. \citet{Suzgun2022ChallengingBT} state, “Codex, trained on both code and text data, shows better performance in following task instructions and exploiting algorithmic patterns based on the prompt exemplars.” \citet{Liang2023HolisticEO} state, “for reasoning-intensive scenarios, we find that the code models, especially Codex davinci v2, consistently outperform the text models, even on synthetic reasoning scenarios posed in natural language.” \citet{Hendy2023HowGA} state that “We hypothesize that the models acquire their reasoning capabilities through training on natural language multilingual data along with programming languages data”.
In summary, these works provide evidence that training LLMs on code serves as indirect supervision for complex reasoning tasks. One of our major goals is to explore the extent to which this hypothesis holds beyond English.

\subsection{Code Prompts for Complex Reasoning}
\label{sec:prior_code_prompts}
Reasoning tasks posed in natural language can be reformulated as \textbf{code prompts}. Using these code-like structures to interact with code-LLMs better aligns the representations seen at training time with those at inference time.  \citet{madaan-etal-2022-language} use few-shot prompting on the Codex LLM to convert tasks into Python graphs, deal with structured commonsense tasks. \citet{zhang-etal-2023-causal} proceed similarly, but for causal reasoning tasks. \citet{chen2023program} consider arithmetic reasoning tasks, and then execute the LLM-generated code on an external interpreter.
The reformulation process from natural language specification to code prompts is an open-ended one, requiring manual annotation effort, creativity, and trial and error.

While these works use code prompts for complex reasoning tasks with classification or numerical outputs, as we did, code prompts can also be applied to tasks with generative outputs, such as knowledge graph construction~\cite{bi2023codekgc} and story understanding~\cite{dong2023corrpus}. To the best of our knowledge, our work is the first to use code prompts in multiple languages.

\subsection{Multilingual Reasoning for LLMs}
The MEGA benchmark~\cite{ahuja2023mega} covers 70 languages and 16 tasks. MEGA considers only simple reasoning tasks, which, as discussed earlier, limits our understanding of how well LLMs can reason across languages.

MGSM \cite{shi2022language} is an arithmetic reasoning dataset in 10 languages, translated from GSM8K~\cite{cobbe2021training}.They find that the chain-of-thought technique (CoT)~\cite{wei2022chain}, by adding to the prompt few-shot examples of step-by-step reasoning, is also effective in the multilingual setting. Interestingly, they find that for non-English questions, English CoT outperforms native language CoT. They further emphasize the reasoning ability increases with model scale. Our xSTREET benchmark is a more comprehensive view of multilingual complex reasoning. xSTREET covers not only arithmetic,\footnote{Instead of using MGSM, we perform our own translation of GSM8k given the intermediate reasoning annotations inherited from~\citet{ribeiro2022street}.}  but adds logic and science tasks, has many more entries, and has ground-truth structured reasoning annotations.

\subsection{STREET Complex Reasoning Benchmark}
\label{sec:street}
The STREET  benchmark is a composite of several complex reasoning tasks~\cite{ribeiro2022street}. The work adds expert human annotations for multi-premise, multistep explanations. Each task's explanation is structured in a reasoning graph. Reasoning graphs, as shown in Figure~\ref{fig:street_task}, consist of nodes which contain statements, and edges that connect nodes. 

\paragraph{Source Tasks} The tasks\footnote{
STREET includes a fifth task, SCONE, which is omitted from xSTREET. SCONE is quite abstract, and involves state-tracking in a toy world. This requires careful consideration beyond translation, and is thus left to future work.} and answer formats are:
\begin{itemize}[noitemsep,topsep=0pt]
    \item \textbf{ARC} science commonsense questions (multiple-choice)
    \item \textbf{GSM8k} arithmetic word problems (number)
    \item \textbf{AQUA\_RAT} arithmetic word problems (multiple-choice)
    \item \textbf{AR\_LSAT} logic problems from a standardized test (multiple-choice)
\end{itemize}

\paragraph{Linearized prompt format}
While a reasoning graph is abstract, to interface with an LLM, \citet{ribeiro2022street} use \textbf{linearized} prompts. This represents a graph as a sequence of tokens, as shown in Figure~\ref{fig:prompt formats}. Statements are given a number index; output statements (i.e., reasoning steps) include a trace of the nodes leading to the new statement. 

Problems with the linearized format arise in that it is task-specific, and that it diverges from LLM's training data distribution. While in-context learning can help the model pattern-match the output format, the underlying reasoning abilities of the LLM may not be properly elicited. Following~\citet{madaan-etal-2022-language}, we argue that interfacing with a code-LLM through code prompts is a more “intuitive” way for the LLM to reason through a task, leading to our novel code prompts format in \S\ref{sec:code_prompts}.

\subsection{Source Code Dataset}
The Stack is a 3.1 TB dataset of permissively licensed source code in 30 programming languages~\cite{kocetkov2022stack}. In this work, we utilize the official small subset\footnote{\href{https://huggingface.co/datasets/bigcode/the-stack-smol}{Available here}}, and consider only 3 popular programming languages: {Java, JavaScript, Python} (10k files each, 30k total).

\section{Multilingual Complex Reasoning Benchmark: xSTREET}
We create the xSTREET dataset by translating STREET into 5 languages: Arabic (ar), Spanish (es), Russian (ru), Chinese (zh), and Japanese (ja). These languages have linguistic and script diversity; furthermore, they are the languages used in many online programming help websites.

To create the \textit{xSTREET test split}, we hire expert human translators for all 5 languages through an internal team (detailed in \S\ref{sec:data_statement}). Translators are tasked with post-editing the machine translation of one sentence at a time; for context, they can refer to the entire STREET entry the sentence comes from. After receiving the translations, we re-use the reasoning graph edges, and replace English nodes with the translations to create xSTREET. This process is shown in Figure~\ref{fig:street_task}. We therefore extend the 914 English entries in STREET to 5484 examples in xSTREET (914 * 6 languages).

To create the \textit{xSTREET train and development splits}, we use machine translation.\footnote{We used an online translation API (anonymized here).} We then asked native speakers to evaluate the quality of 10 random sampled translations of each language. Annotators gave feedback that, despite some errors, the translations were of reasonable enough quality to use for training purposes.

Dataset statistics for the xSTREET test benchmark are given in Table~\ref{tab:xstreet_statistics}.
\begin{table}[!t]
\small
\centering
\begin{tabular}{@{}llll@{}}
\toprule
Dataset & \begin{tabular}[c]{@{}l@{}}\# entry \\ /lang\end{tabular} & \begin{tabular}[c]{@{}l@{}}\# sents/\\ lang\end{tabular} & \begin{tabular}[c]{@{}l@{}}avg \# \\ sents/entry\end{tabular} \\ \midrule
ARC & 340 & 4334 & 12.7 \\
AQUA RAT & 254 & 3436 & 13.5 \\
AR LSAT & 50 & 1158 & 23.2 \\
GSM8k & 270 & 2255 & 8.4 \\ \midrule
\multicolumn{1}{r}{Total} & 914 & 11183 & 12.2 \\
\multicolumn{1}{r}{x6 languages} & 5484 & 67098 &  \\ \bottomrule
\end{tabular}
\caption{Statistics for the xSTREET test benchmark. }
\label{tab:xstreet_statistics}
\end{table}

\section{Code with Multilingual Comments as Indirect Supervision for Reasoning}
Taking the idea of using code for reasoning, and comments for multilinguality a step further, we address the question: \textit{can multilingual code serve as indirect supervision for multilingual reasoning?} In other words, we investigate whether the code \& reasoning hypothesis holds multilingually. We therefore propose a lightweight fine-tuning recipe, which consists of creating a multilingually commented code dataset, then fine-tuning on it, which serves as \textit{indirect supervision} for downstream reasoning tasks.

\subsection{Translated Code Comments Dataset (\tcc)}

The first step of the recipe is creating a source code dataset with \underline{t}ranslated \underline{c}ode \underline{c}omments, termed \tcc. For each file from the source dataset, and for each target language, we perform the following. We parse the code to extract out comments, translate comments into the target language, then replace the original comments with translations. This is depicted in Appendix Figure~\ref{fig:tcc_creation}.

We use two simple filters: for source code files that A) have >5 comments, and B) whose comments are over 50\% in English.\footnote{We performed other filtering experiments, described in Appendix~\ref{sec:tcc_filtering}, which had similar performance.} This filters 30k source code files down to 20k.
After translating into 5 additional languages, TCC consists of 20k*6=120k files total.
See Appendix Table~\ref{tab:tcc_statistics} for dataset statistics.

\subsection{Train Time: fine-tuning on \tcc}
\label{sec:ft}
In the second step, we leverage low-rank adaptation (LoRA)~\cite{hu2021lora} to finetune instruction-tuned LLMs on \tcc.\footnote{We used a g5.48xlarge instance from AWS, which has 8 NVIDIA A10G GPUs (24*8 GB=192GB vRAM).} We use two methods to preserve the original model's capabilities despite the additional finetuning. First is by using LoRA itself, as it keeps the original base model's parameters frozen and introduces only a few learned parameters. Secondly, we replay 100k examples from the base model's training data, xP3~\cite{muennighoff2022crosslingual}, in a multitask setup with the \tcc\ LM task.

The recipe for a reasoning-enhanced LLM is now complete, and this is depicted in Figure~\ref{fig:recipe}. 

\section{Multilingual Complex Reasoning as a Downstream Task}
\label{sec:code_prompts}

We hypothesize that structure, when applied to reasoning problems formulated in different languages, can abstract away some of the language-specific details, better surfacing the reasoning steps needed for a model. We thus propose the \Sim\ (\underline{S}elect-and-\underline{i}nfer \underline{m}ultilingual comments) code prompts for complex reasoning tasks.

\Sim\ code prompts utilize several functions. We do not provide the API definitions, instead, we expect the model to learn to use them from the in-context examples.  The functions are:
\begin{itemize}[noitemsep,topsep=0pt]
    \item \texttt{select\_facts(facts)}
    \item \texttt{infer\_new\_fact(selected)}
    \item \texttt{is\_solved(fact, question)}
    \item \texttt{make\_choice(fact, choices)}\footnote{This function is not used for non-MC tasks, i.e. GSM8k.}
    \item \texttt{facts.append(fact)}
\end{itemize}

\texttt{select\_facts} and \texttt{infer\_new\_fact} are loosely inspired by Selection-Inference~\cite{creswellSelectionInferenceExploitingLarge2023}. A key difference, though, is that we use a single prompt, instead of iterative prompts. We therefore include \texttt{is\_solved(fact, question)} as a signal for the LLM to stop generation.

Each function is annotated with its return value in an inline code comment. This is inspired by prior work~\cite{zhang-etal-2023-causal}. \texttt{infer\_new\_fact} has a string return value, i.e., the text of the new fact. We experiment with two versions of the return value of \texttt{select\_facts}. The first, termed \Sim-indexed, uses variables \texttt{facts[i]} to reference the \texttt{facts} array (similar to the indices used in linearized format). The second, termed \Sim-text, directly uses each fact's text, dereferenced from \texttt{facts[i]}. We find that \Sim-text works best for smaller models, while \Sim-indexed does for larger ones, and hence apply this going forward.

\begin{figure*}[t!]
    \centering
    \includegraphics[width=.8\textwidth]{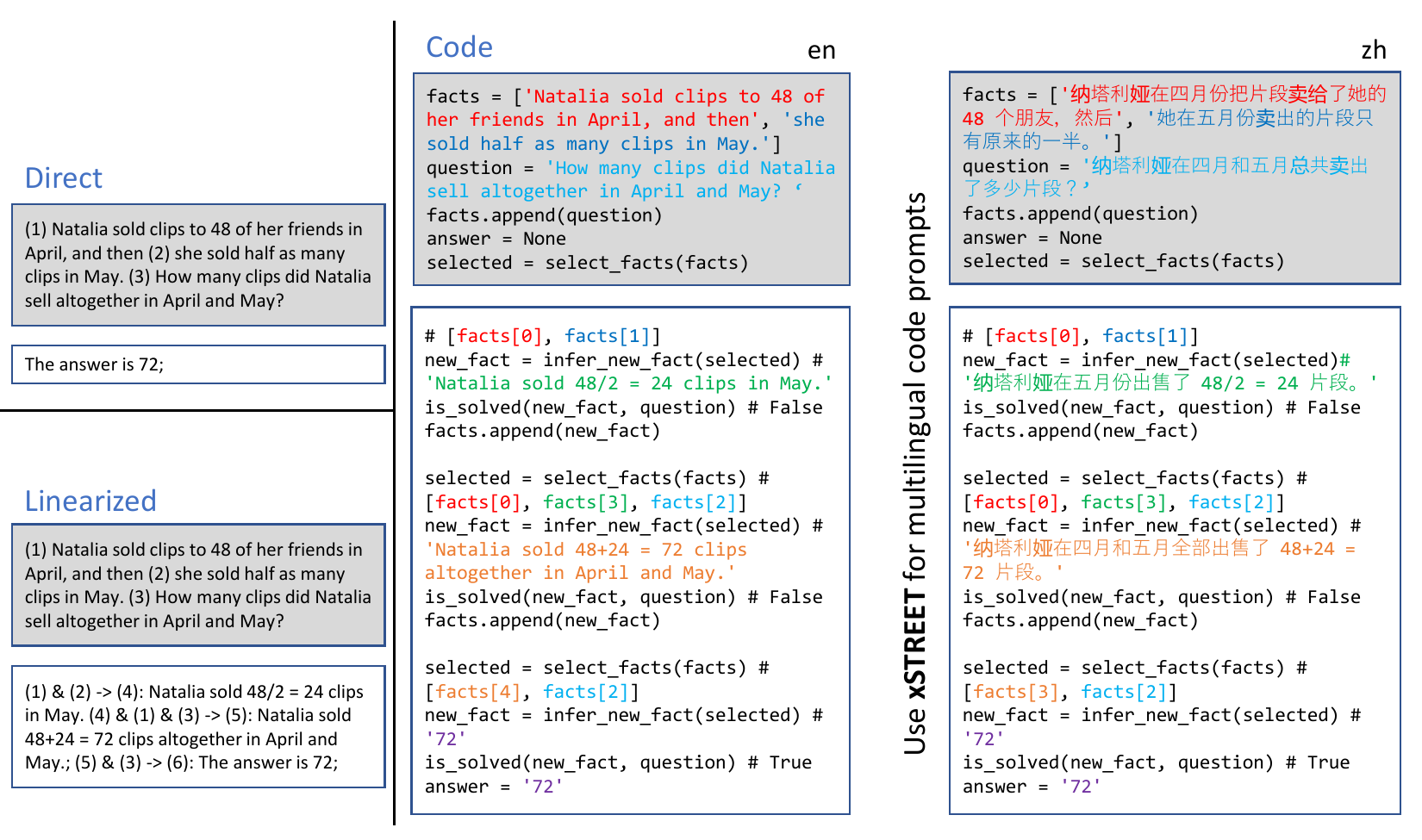}
    \caption{Depictions of 3 prompting formats for the xSTREET tasks. For each format, input is in a grey box, while expected output is in a white box. Top left: direct. Bottom left: linearized. Right: \Sim\ code prompts (2 languages). In the code prompts, we color code facts which are aligned.}
    \label{fig:prompt formats}
\end{figure*}

We write a rule-based Python script that converts existing structured graph annotations to \Sim\ code prompts. \Sim\ prompts express the exact same information as the linearized format. This property is unlike code prompts for prior work, wherein the conversion is done through in-context learning with an LLM, which can introduce errors as discussed in~\S\ref{sec:prior_code_prompts}. The different prompting formats for LLMs are shown in Figure~\ref{fig:prompt formats}.
\paragraph{Multilingual code prompts} We use multilingual input in \Sim~code prompts as follows. First, facts given in the question are listed in the language of the task in a list of strings. Second, new facts and selected facts are given as comment lines adjacent to the function calls. See Figure~\ref{fig:prompt formats} for an example.

\section{Experimental Setup}

\paragraph{Models Used} We primarily study open-source models, which allows for application of both train and inference-time techniques. We use \textbf{BLOOMZ}~\cite{muennighoff2022crosslingual} as our base LLM. This model is instruction-finetuned on prompts in 46 natural languages and 10 programming languages. For our experiments, we consider the 7.1B-parameter BLOOMZ, as well as BLOOMZ-\tcc\, which is further finetuned on \tcc.

For inference-time only, we consider two larger LLMs. We use the instruction-finetuned version of \textbf{Falcon}~\cite{almazrouei2023falcon} (40B), another open-source LLM trained on text+code. Compared to BLOOMZ, Falcon is more performant on English tasks; however, it has limited multilingual abilities.\footnote{As stated in the \href{https://huggingface.co/tiiuae/falcon-40b-instruct}{model card} for falcon-40b-instruct.}  We also use \textbf{GPT-3} (175B)\footnote{\texttt{text-davinci-002} following~\citet{ribeiro2022street}}, a closed-source model that is popularly-used and powerful.

\paragraph{Prompting setup} We use few-shot prompting, and random sample up to 5 exemplars from the train split (up to a model's context length).\footnote{Due to brevity, Figure~\ref{fig:recipe} uses a 0-shot prompt, and only depicts the \Sim\ prompting format. The reported results use 5-shot prompts, and are given for all prompting formats.} For each inference example, the same exemplars are used for all models and prompt types. We use greedy decoding, and task the model with generating up to 682 tokens. (max context length of BLOOMZ 2048 // 3).\footnote{We acknowledge a limitation with the max of 682 tokens, as this will truncate output for questions which require longer reasoning chains.}

\section{Results}
We report results on the xSTREET benchmark. We use the answer accuracy metric, adapting evaluation from~\citet{ribeiro2022street}.\footnote{\label{graph_sim}While STREET also measure graph similarity between linearized output and reference graphs, we did not implement them for \Sim\ prompts. This is because for the small LLMs (7B), even the linearized format had near 0 graph similarity.}

Given the extensive nature of the xSTREET benchmark and our model experimentation, we highlight our findings iteratively. We first consider only BLOOMZ and BLOOMZ-\tcc, with a particular focus on ARC,  where our methods are the most impactful.  We then consider GPT-3 and Falcon.

The full results are given in Appendix~\ref{sec:full_results}. Here we provide numbers for all tasks, languages, models, and prompt formats (this also includes the direct prompting format).

\subsection{Results for BLOOMZ Models}

\begin{figure*}[ht!]
    \centering
    \includegraphics[width=.9\textwidth]{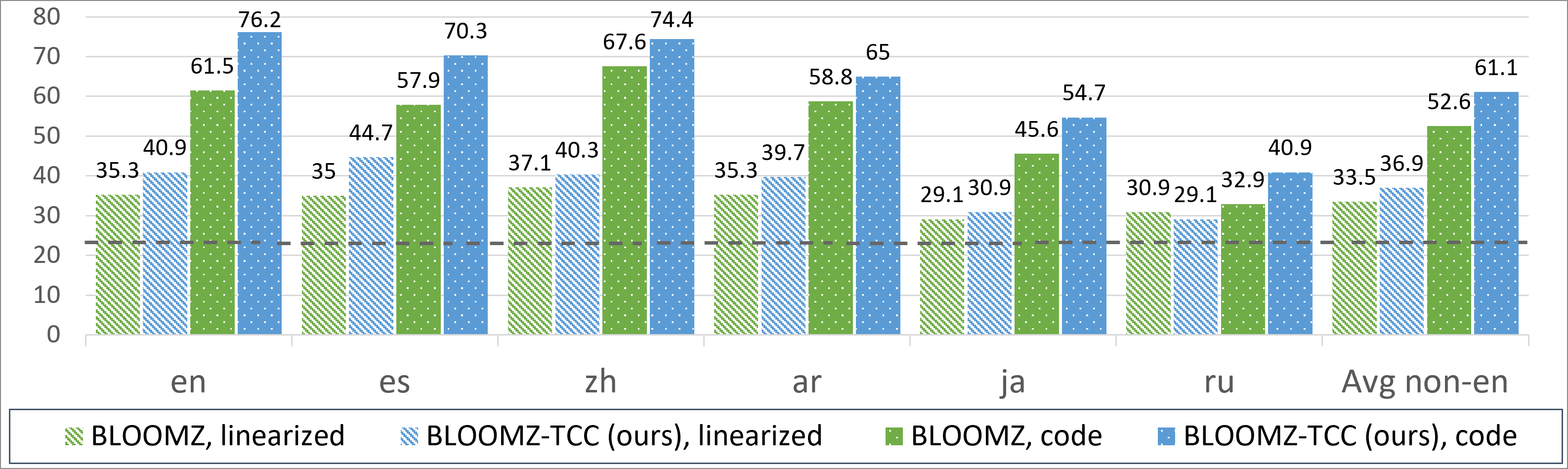}
    \caption{Results on ARC task of xSTREET, with BLOOMZ-based models. The random baseline is 25\%. `Avg' bars are across the 5 non-English languages. Linearized prompts use lines, while code prompts use dots. }
    \label{fig:arc_results}
\end{figure*}

Results for the ARC task (science commonsense reasoning) are shown in Figure~\ref{fig:arc_results}. Several takeaways arise. We see that code prompts greatly outperform linearized prompts across all languages.  For results within a single model, reasoning performance drops greatly comparing English vs. average non-English (e.g.~from  76.2\% to 61.1\% accuracy for BLOOMZ-\tcc). This provides evidence that current multilingual LLMs are still optimized for English. This underscores the usefulness of xSTREET for developing LLMs with better underlying, language-agnostic abilities.

We next turn to comparing base BLOOMZ vs. our finetuned BLOOMZ-\tcc. We see that BLOOMZ-\tcc\ outperforms BLOOMZ for all languages and both formats. More interestingly, relative multilingual gain is much larger when using code prompts vs.~linearized prompts (Avg non-en, 52.6 $\rightarrow$ 61.1 vs. 33.5 $\rightarrow$ 36.9). This is evidence that the code prompt format improves multilingual reasoning, likely by the explicit separation of the reasoning task (in code) vs the multilingual understanding (in comments).
Finally, looking at per-language trends for BLOOMZ-\tcc\, we see that code prompts are most effective for en, es, zh, and ar, while less so for ja and ru.\footnote{This is likely because the base model, BLOOM, was not trained on any ja or ru text.}

\paragraph{Results on GSM8K, AQUA\_RAT, AR\_LSAT}

Our results show that BLOOMZ and BLOOMZ-\tcc\ struggle for the other tasks, with performance being around random chance whether using the interventions or not. 
We hypothesize that these tasks are ``too hard'' for the BLOOMZ-7B used; to reiterate, GSM8K and AQUA\_RAT are arithmetic reasoning, while AR\_LSAT is logical reasoning.  This concurs with the common view that complex reasoning capabilities of LLMs are emergent with model scale~\cite{wei2022emergent}.  We further discuss these results, and expand our hypotheses, in Appendix \S\ref{sec:bloomz_other_results}.

\subsection{Results for Larger LLMs}
\label{sec:gpt_results}
\begin{figure*}[t!]
    \centering
    \includegraphics[width=.9\textwidth]{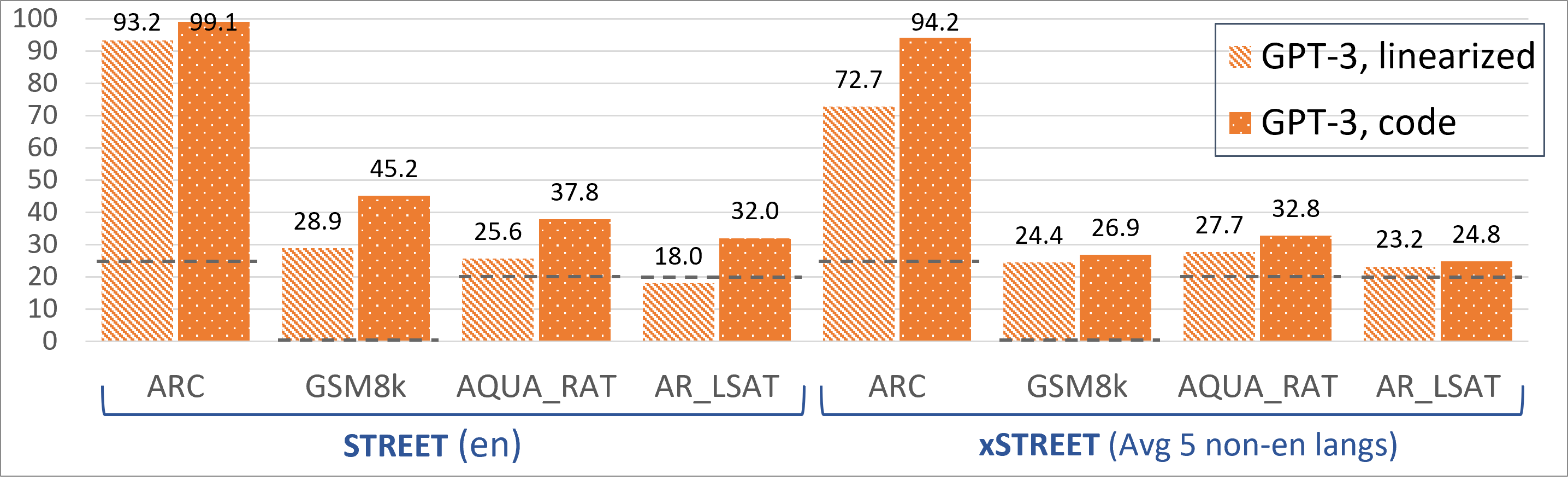}
    \caption{Results on GSM8k, AQUA\_RAT, AR\_LSAT tasks of STREET (left) and xSTREET (right), with GPT-3. For each task, the random baseline is shown with a dotted line. xSTREET results are averaged over 5 languages.}
    \label{fig:gpt_results}
\end{figure*}

As code prompts are at inference time, they can be used to interface with any LLM. We report results for GPT-3 in Figure~\ref{fig:gpt_results}. We see as before that the multilingual setting poses additional challenges for reasoning, as English results are always higher than corresponding non-English tasks.

First considering ARC, GPT-3 performs strongly in English for both formats, nearly solving the task. Comparing English to multilingual ARC, linearized suffers a sharp drop (93.2 $\rightarrow$ 73.2), while code prompts remain robust (99.1 $\rightarrow$ 94.2). This underscores the effectiveness of \Sim\ prompts in disentangling the reasoning and multilingual components of the task.

For the other tasks, \Sim\ always outperforms linearized format. Comparing relative gains, code prompts boost performance more in English than on multilingual settings. While still a very positive result, this differs from ARC as discussed above. To discuss why this is the case, we consider the dual effects of \Sim\ code prompts, vs. linearized: the function calls capture the reasoning structure, while the multilingual comments capture the language understanding. Because the arithmetic and logical reasoning tasks are far more symbolic than the ARC commonsense reasoning task, multilingual language understanding is less effective.

\begin{table}[t!]
\small \centering
\begin{tabular}{@{}llll@{}}
\toprule
Model & XNLI & \begin{tabular}[c]{@{}l@{}}XStory-\\Cloze \end{tabular} & XQUAD \\ \midrule
BLOOMZ & 45.5 & 72.4 & 80.5 \\
BLOOMZ-\tcc\ (ours) & 45.6 & 71.8 & 80.4 \\ \bottomrule
\end{tabular}
\caption{Results for 3 non-complex multilingual reasoning tasks, averaged over all languages.
}
\label{tab:xnli_results}
\end{table}

\subsection{Non-Complex Reasoning Task Results}
Recall that our fine-tuning recipe aims to improve reasoning of an LLM, while maintaining its natural language understanding (NLU) abilities. We show this is the case by reporting results on 3 multilingual tasks: 
\begin{itemize}[noitemsep,topsep=0pt]
\item \textbf{XNLI}: natural language inference 
\item \textbf{XStoryCloze}: given 4 sentences from a short story, choose between 2 possible completions 
\item \textbf{XQUAD}: extractive question answering
\end{itemize}
To query LLMs, we follow the specific prompting guidelines for each task from~\citet{ahuja2023mega}. Table~\ref{tab:xnli_results} shows that for all 3 tasks, the differences between  BLOOMZ and BLOOMZ-\tcc\ are statistically insignificant. Therefore, the mitigation strategies we used, LoRA and training data replay, have proved effective.

\begin{figure*}[t]
    \centering
    \includegraphics[width=.9\textwidth]{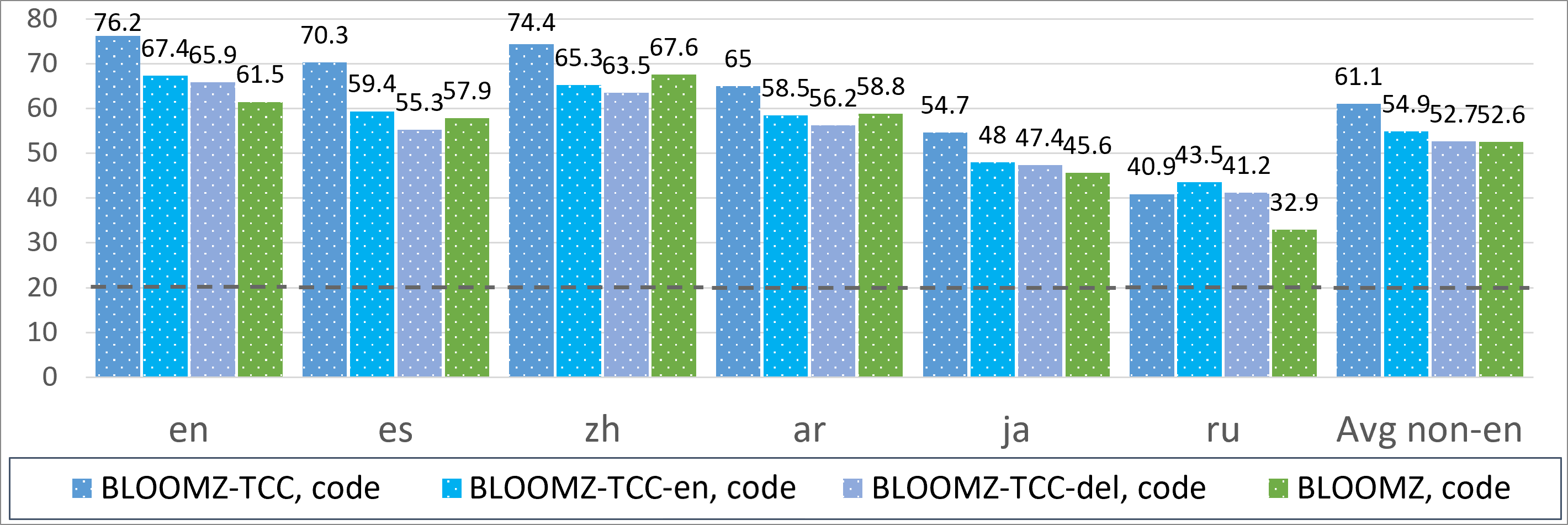}
    \caption{Results on ARC subtask of xSTREET for the code ablation experiments, where BLOOMZ is finetuned on different datasets. 
    BLOOMZ-\tcc\ uses our proposed multilignual code comment augmentation process, BLOOMZ-\tcc-en uses the source code files with English comments, and BLOOMZ-\tcc-del uses source code files with all comments deleted. 
    As in Figure~\ref{fig:arc_results}, we use \Sim\ prompts with up to 5-shot examples, and show `Avg' bars across the 5 non-English languages. }
    \label{fig:ablation}
\end{figure*}

\subsection{Effect of Code Comments on Downstream Reasoning}
The code \& reasoning hypothesis speaks to training on code improving LLM reasoning. However, an integral part of source code is comments, which have been underexplored by prior work. We study 2 ablation settings, with the same finetuning setup:  \textbf{\tcc-en}: original source code files (i.e. English-only comments). \textbf{\tcc-del}: source code files without any comments (comments are deleted).

We evaluate the best prompt format (\Sim) on the ARC subtask. Results are shown in Figure~\ref{fig:ablation}. We see that overall, finetuning on \tcc\ is the best configuration, then \tcc-en, and finally \tcc-del. These trends generally hold over the 6 languages.\footnote{Russian (ru) is an exception, where BLOOMZ-\tcc-en outperforms BLOOMZ-\tcc. We will investigate this further, but note this may be an artifact of the base LLM, BLOOM, not having tokenization for Cyrillic script.}

This ablation study adds a new consideration to the code \& reasoning hypothesis: that within code, even the comments are influential in downstream reasoning performance. Furthermore, we see that the diversity of code comments introduced by our proposed data augmentation of \tcc\, further boosts performance in all languages, including English.

\section{Analysis}
To further understand wherein our techniques help, or fail to help, model reasoning, we perform some manual analysis. For brevity, we focus on 2 languages, en and ar, and 2 tasks, ARC and GSM8K. We first perform error analysis on BLOOMZ, then perform a case study for each task.

We further perform 3 additional experiments, which are detailed in Appendix \S\ref{sec:ablations}. To highlight one interesting finding, we show that training on diverse code comments, such as from the multilingual \tcc, boosts xSTREET performance in all languages including English. 

\paragraph{Error Analysis for BLOOMZ English}%
\label{sec:arc_qa}
For this task, and with base BLOOMZ, \Sim\ achieves 61.5 ARC accuracy, while linearized achieves 35.3. Our manual analysis of outputs reveals that the performance discrepancy is largely due to \textit{poor instruction-following when using linearized} vs. using \Sim.  \citet{ribeiro2022street}  find that for linearized (and their model), 62\% of generations fail to generate a parsable answer (i.e., reasoning graph is incomplete). Our findings concur, in that linearized has 66\% (223/340) invalid generations. In contrast, \Sim\ has only \~{}19\% invalid. BLOOMZ-\tcc\ with \Sim\ further reduces invalid rate to 9\%, and increases accuracy to 76.2. 
We observe that in cases where all formats output successfully, the reasoning graph and answers are nearly identical. The difference is that \Sim\ prompts allows the model to generate a complete reasoning graph far more often. We reiterate that this behavior is a novel finding given BLOOMZ-\tcc\ was indirectly supervised on code, rather than directly on reasoning tasks. Further discussion is found in Appendix~\ref{sec:discussion}.

We summarize this section with the following view: our techniques \textit{elicit better instruction-following} of the proscribed reasoning format from a base LLM, leading to improved benchmark performance. Within a reasoning step, the models are making similar decisions, but at the reasoning-graph level, our methods assist in harder cases.
 
\subsection{Case Study on GSM8k English}
We perform a case study of one GSM8k problem, comparing 3 models (BLOOMZ, BLOOMZ-\tcc, GPT-3) and 2 formats (linearized, \Sim) in Appendix Table~\ref{tab:gsm8k_en_study}. We observe that only GPT-3 with \Sim\ achieves the correct answer, and reasoning steps also concur with the gold completion. GPT-3 with linearized representation makes an erroneous first step, which propagates the error downwards. Both BLOOMZ models with linearized formatting only follow the output format, and the text statements are copied from the input instead of being new statements. BLOOMZ with \Sim\ has repetitive output and does not output an answer. While BLOOMZ-\tcc\ still outputs a wrong answer, it does perform 2 rounds of reasoning through selecting and inferring facts. So, we see that both interventions elicit better underlying reasoning abilities of LLMs.

\subsection{Case Study on ARC Arabic}
We look at an Arabic example from ARC in Appendix Table~\ref{tab:arc_ar_study}. We observe that for the linearized format, the final answer is incorrect (A), given the model makes a wrong penultimate inference. The \Sim\ format, meanwhile, allows GPT-3 to output the correct answer (D), given it makes a correct inference step (albeit 1 step less than the gold). In fact, directly prompting GPT-3 leads to a correct answer.
This again highlights the importance of aligning the prompt format, which is code here, to the training format.

\section{Conclusion}
 We introduced xSTREET, a multilingual structured reasoning benchmark which covers 5 diverse languages 
 , spans science commonsense, arithmetic and logical reasoning tasks, and includes high-quality intermediate reasoning steps. We found that current multilingual LLMs underperform in the non-English setting, then proposed two methods to remedy this, based on the popular hypothesis that LLMs trained on code are better reasoners. At training, we propose translating the comments of a source code dataset, to use as indirect supervision data for parameter-efficient fine-tuning. During inference, we leverage code structure to represent reasoning graphs. We perform extensive experimentation, and both of our methods better elicit underlying reasoning abilities of LLMs. 

Our work brings together two areas of challenge for LLMs — multilinguality, and complex reasoning. In particular, our fine-tuning recipe shows that the code \& reasoning hypothesis can apply multilingually. We suspect that improvements can be amplified if multilingual comments are included at the pre-training, instead of the fine-tuning stage. We hope our findings underscore the key role that code should play in the development of LLMs with better reasoning capabilities across languages.

\section*{Limitations}
One limitation is that we were unable to apply our fine-tuning recipe to the stronger LLMs. ``Stronger'' refers to two characteristics. First and unavoidably, we can only apply the method to weaker open-source models, as closed-source models are proprietary; nevertheless, we explored them with our inference-time \Sim\ prompts approach, and this worked well. Second, we only were able to fine-tune a 7B parameter model due to our resource constraints, so it is to-be-determined the effectiveness of the recipe on 70B+ models.

Between the submission and publication of this work (February to August 2024), LLM development has been brisk, and several recently released $\sim$7B LLMs have shown decent performance on arithmetic reasoning. In our work, we were limited to BLOOMZ-7B, which we saw was poor at math. For followup work, therefore, we are excited to try our finetuning approach on \tcc\ while using these newer LLMs as base models.

Another limitation is for the xSTREET benchmark, we performed human translation on only the test set of the source STREET dataset. As we used machine translation for the train set, but also drew few-shot exemplars from these, the lower exemplar quality worsens performance compared to a gold standard exemplars. We also fine-tuned on machine-translated TCC.

While we tried to be inclusive with the languages chosen, studying 6 languages from different families and using different scripts, we acknowledge that more community effort will need to go into expanding the study of multilingual complex reasoning to lower-resource languages. We further acknowledge the limits of the translation of English reasoning tasks and intermediate steps alone, in that reasoning processes may differ for speakers of different languages. So too may a multilingual LLM respond inconsistently to queries posted in different languages~\cite{li2024land}, which warrants future studies into how this holds for the reasoning tasks studied in this work.

Finally, in this work, we considered only the final answer accuracy for the tasks. The original STREET tasks from \citet{ribeiro2022street} included various graph similarity metrics used to consider the intermediate reasoning steps as well -- a definite strength of their structured reasoning approach vs. unstructured approaches such as CoT. We did not do this consideration due to the difficulty of reimplementing the graph similarity metric calculation for the different languages, and leave this to follow up work. Furthermore, we note that the 7B LLM we used had overall poor graph similarity (near 0 for all metrics) using the original STREET evaluation scripts and dataset. 

\paragraph{Data Statement} \label{sec:data_statement}  We provide a data statement in adherence with the ACL code of conduct and recommendations laid out in Bender and Friedman (2018). Linguists working on the Machine Translation Post Editing project for the multilingual dataset into Arabic, Chinese, Japanese, Russian, and Spanish are in-country, native speakers. They all are certified translators with more than 5 years of full-time translation experience, according to the 17100 Translation ISO Standard. These linguists were hired through vendors and were remunerated above industry standard rates. Instructions were to post-edit machine translated output and included guidelines on what to localize (artist names, city names, metric conversions), format (capitalization, punctuation) and structure (sentence level breaks). The vendor project managers made sure the instructions were adhered to. The QA process consisted of content review based on the Multidimensional Quality Metric (MQM) model that allocates different weights to 5 error severities (0-none to 5-critical) in several error topics. Total sample reviewed was 5 (5k words) of the total (100k words) source word count.

 \section{Acknowledgements}

 We would like to thank Danilo Neves Ribeiro for his guidance on working with the STREET benchmark, and insightful conversations on how tackle our multilingual extension of complex reasoning. We thank several colleagues for providing annotations: Etsuko Ishii, Igor Shalyminov, Yuwei Zhang. We thank these people for discussion and feedback: Salvatore Romeo, Yi Zhang, Sam Davidson, and Sailik Sengupta.

\bibliography{anthology,custom}
\bibliographystyle{acl_natbib}

\appendix
\input{appendix.tex}

\end{document}

%% file: appendix.tex
\input{gsm8k_en_ex}
\input{arc_arabic}
\begin{figure}[t!]
    \centering
    \includegraphics[width=.9\linewidth]{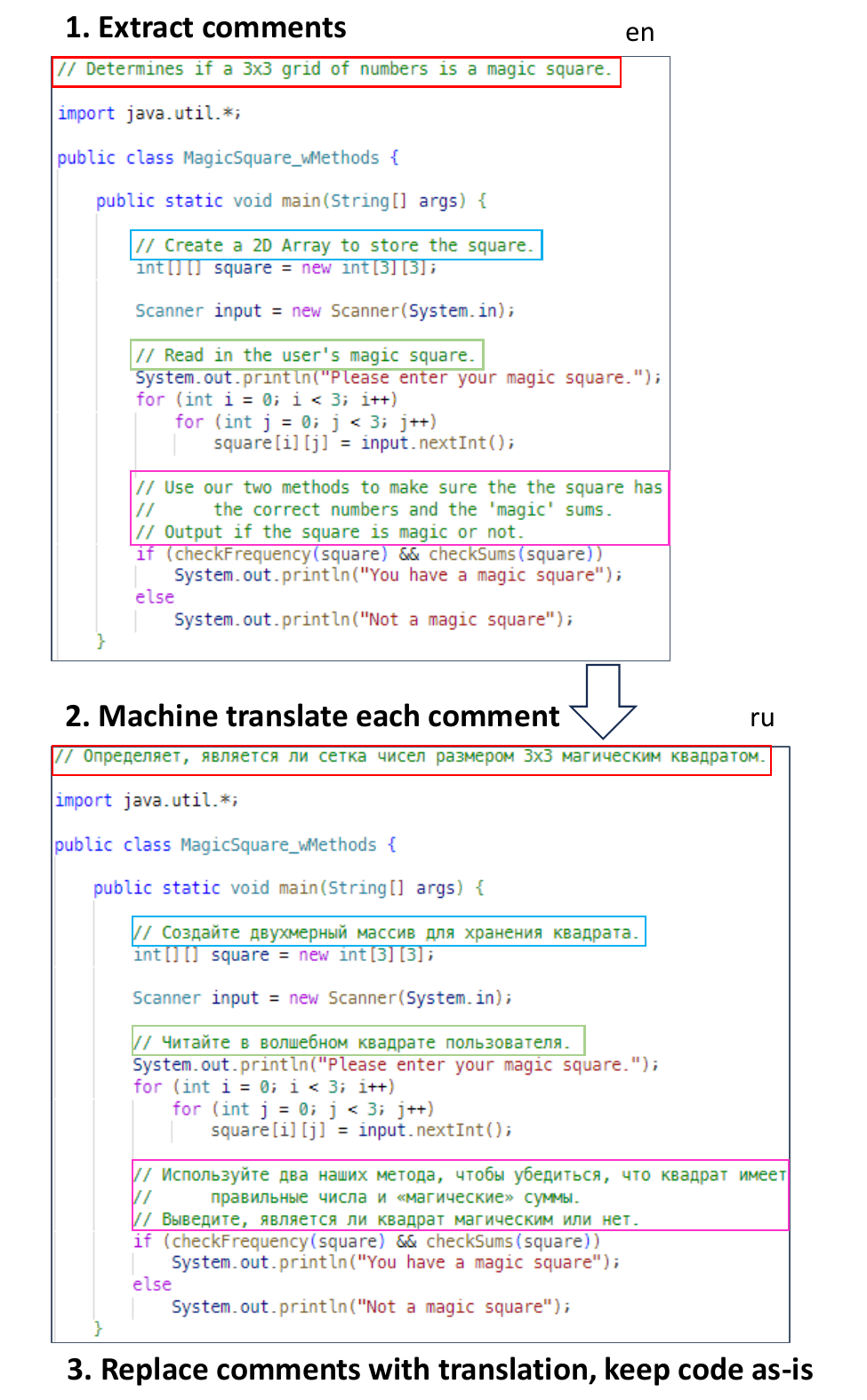}
    \caption{The data augmentation approach used to create the \tcc\ (translated code comments) dataset.}
    \label{fig:tcc_creation}
\end{figure}

\section{Further Discussion}
\label{sec:discussion}

We find that regardless of answer correctness, BLOOMZ-based models often fail to generate new text, instead of copying text from the input. This again is likely due to the weaknesses of BLOOMZ, as this is not observed for GPT-3 with any format. Our use of both interventions, greatly reduces the incidence of this problem, which as we have discussed leads to BLOOMZ-\tcc better eliciting the model's underlying reasoning abilities.

\subsection{BLOOM Results for GSM8K, AQUA\_RAT, AR\_LSAT}
\label{sec:bloomz_other_results} 
These results are shown in Appendix Figure~\ref{fig:other_results}. 

For all tasks, performance is around random chance. For GSM8K, random chance is 0, and the models fails to solve nearly any math problem. While all numbers are close and likely statistically insignificant, we see that BLOOMZ-\tcc\ slightly underperforms base BLOOMZ, and linearized and code prompts perform similarly. 

Our hypothesis on why this happens, as discussed before, builds on the view that truly complex reasoning capabilities are emergent with LLM's model scale. The 7B BLOOMZ model used has no baseline ability for these 3 tasks (while it did for ARC), and therefore our interventions, which are indirect supervision on code, cannot help elicit better reasoning. 

We discuss the two interventions separately. First, we study the effectiveness of code prompts on larger LLMs in \S\ref{sec:gpt_results}. Second, for the finetuning recipe, we draw some initial points in Appendix~\ref{sec:discussion}, given our resource constraints on small LLMs.\footnote{With 192 GB vRAM, we could finetune at most 7B multilingual models (which have much larger vocabulary sizes and thus larger embeddings). We leave future work to use our recipe with larger models, such as by using 4-bit quantization.} 

This suggests limitations to the code+reasoning hypothesis, which have not been adequately discussed in prior work. Indirectly supervising LLMs for reasoning by training on code is effective for specific types of reasoning, such as ARC's commonsense reasoning, and less so for math problems like GSM8K, though, intuitively, code probably does not help, given code rarely includes arithmetic equations.\footnote{Arithmetic reasoning can be improved by \textit{directly} finetuning on math -- \citet{Ni2022LearningMR} achieve 19.5\% on GSM8k on a 2.7B LLM with this approach.} As for the logical reasoning problems of AR\_LSAT, we defer study to future work applying our finetuning recipe to larger LLMs. 

\section{Hyperparameters}
For the TCC finetuning recipe, we set maximum sequence length to 1024, set learning rate to 1e-5 (with 0.1 weight decay), do not use warm-up, and use cosine learning rate schedule. We trained for 2 epochs using a batch size of 3 and gradient accumulation over 20 steps. We set the LoRA layers dimension to 128. The implementation is done with the DeepSpeed-Chat framework \cite{yao2023dschat} and the \texttt{transformers} library \cite{wolf-etal-2020-transformers}.

\section{Dataset Statistics}
Statistics for TCC are shown in Appendix Table~\ref{tab:tcc_statistics}.

\begin{table}[ht!]
\centering
\begin{tabular}{@{}ll@{}}
\toprule
\# files from source & 30000 \\
TCC \# files/lang & 20289 \\
\multicolumn{1}{r}{x6 langs} & 121734 \\
\# tokens per language & 55-60m \\ \bottomrule
\end{tabular}
\caption{Statistics for the \tcc\ dataset. The source dataset is the small \href{https://huggingface.co/datasets/bigcode/the-stack-smol}{subset} of The Stack for \{js, py, java\}. The \# of tokens is calculated with the BLOOMZ tokenizer, which has a 250k vocabulary size including tokens from multiple natural and programming languages.}
\label{tab:tcc_statistics}
\end{table}

\section{Full Results}
\label{sec:full_results}
We now report results of all experiments and settings studied in this work. Table~\ref{tab:full results} gives results for all models and prompt formats, on STREET and xSTREET (averaged across 5 languages). Table~\ref{tab:per_lang_results} gives per-language xSTREET results. We first discuss BLOOMZ and GPT-3, which were analyzed in the paper, then separately discuss Falcon.

\subsection{Direct Prompting}
These tables include the direct prompting strategy, in which the model is given the same input as linearized, but needs to generate only the answer without intermediate reasoning. 

For ARC, we find that, surprisingly, direct outperforms linearized (all LLMs). This is likely because the ARC questions are relatively easy already, and directly solving them given the context as well is possible. As discussed earlier, linearized format is artificial and hard to follow, which causes many reasoning graphs without valid answers. In contrast, \Sim\ prompts are captured in code, which the models have seen, and therefore \Sim\ results outperform direct and linearized.

For GSM8k and AQUA\_RAT (using GPT-3), direct prompting fails. Using intermediate reasoning, as found by many prior works, is essential for arithmetic problem-solving. Linearized boosts performance significantly, and \Sim\ code even further.

Regarding xSTREET, overall trends are relatively consistent with those discussed above.

\subsection{Results for Falcon}
Falcon is an open-source LLM that is more performant than BLOOMZ, albeit English-centric. We chose the 40B instruction-finetuned variant, intermediate between BLOOMZ 7B and GPT-3 175B.\footnote{BLOOMZ has 7B and 176B variants, but nothing in between.}

First, we consider STREET results. For ARC, Falcon fails with direct prompts (34.7), but does much better with linearized (76.5) and \Sim\ (81.5). For GSM8k, now that the base model has some math ability with direct prompts (4.4), it can improve with linearized (28.9) and \Sim\ (19.6). \Sim\ underperforming linearized here is because of a context size issue.

Falcon has a max context length of 2048 tokens, while GPT-3 and BLOOMZ can accept up to 4096. \Sim\ code prompts use a lot of tokens for the code structure, and therefore, Falcon will run out of tokens quickly and therefore fail to generate a full reasoning graph in more cases than when using linearized. This is more so a limitation of Falcon than our work (recall that most prior work considers complex reasoning tasks with huge closed-source models).

For AQUA\_RAT, Falcon performance is near random (20) for all 3 prompt formats; i.e., it is ``too hard''. For AR\_LSAT, Falcon is near random (20) for direct and linearized, but achieves 34.0 with \Sim\ prompts.

\paragraph{Multilingual performance}
Even though Falcon is an English-centric LLM, we evaluate its performance on xSTREET. We see that linearized performs the best across all tasks, with code prompts behind, and direct even further behind. Again, we attribute this to Falcon's shorter context length of 2048 -- which is especially non-optimal for 4 of 5 languages studied which do not use the Latin script. The Falcon tokenizer did not see these scripts, resulting in byte-level tokenization, which further uses up the budget.

\begin{figure*}[t]
    \centering
    \includegraphics[width=.95\textwidth]{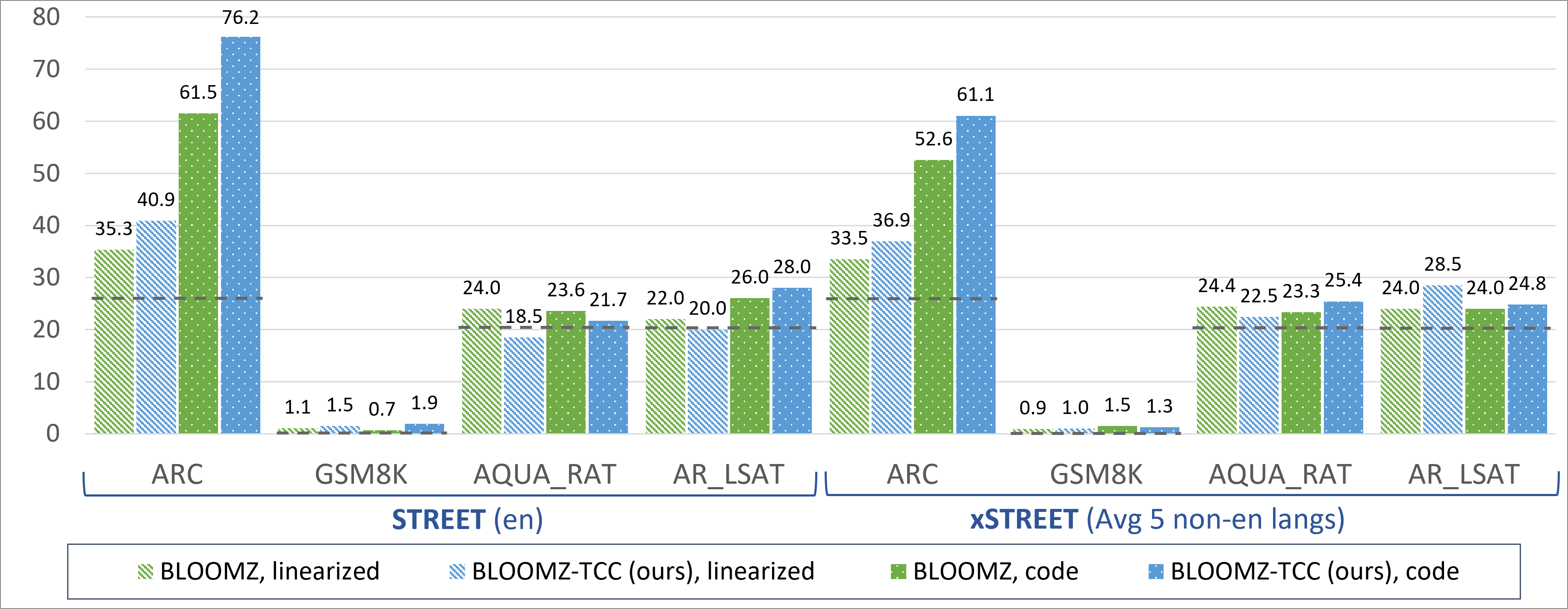}
    \caption{Results on GSM8K, AQUA\_RAT, AR\_LSAT subtasks for BLOOMZ-based models, with (up-to) 5-shot prompts. Random baselines are indicated with dashed grey lines. For each task, we report results 1) for, and 2) averaged over the 5 languages.}
    \label{fig:other_results}
\end{figure*}

\begin{table*}[ht!]
\small
\setlength{\tabcolsep}{4pt}
\begin{tabular}{@{}ll|llll|llll@{}}
\toprule
                            & \multicolumn{1}{r|}{} & \multicolumn{4}{c|}{STREET (English)} & \multicolumn{4}{c}{xSTREET (average 5 languages)} \\ \midrule
                 & Model                 & ARC   & GSM8k  & AQUA\_RAT & AR\_LSAT & ARC      & GSM8k     & AQUA\_RAT    & AR\_LSAT    \\ \midrule
\multirow{4}{*}{\rotatebox{90}{direct}}     & BLOOMZ                & 54.7  & 2.6    & 24.8      & 20.0       & 41.7     & 1.7       & 21.1         & 20.8        \\
                            & BLOOMZ-TCC (ours)    & 70.3  & 3.3    & 20.9      & 24.0       & 49.0     & 1.8       & 21.9         & 23.2        \\
                            & falcon-40b-instruct   & 34.7  & 8.5    & 24.8      & 26.0       & 32.0     & 6.1      & 24.6         & 24.0            \\
                            & GPT-3                 & 97.6  & 4.4    & 21.7      & 22.0       & 90.7     & 1.6       & 25.6             & 24.4             \\ \midrule
\multirow{4}{*}{\rotatebox{90}{linearized}} & BLOOMZ                & 35.3  & 1.1    & 24.0        & 22.0       & 33.5     & 0.9       & 24.4         & 24.0        \\
                            & BLOOMZ -TCC (ours)    & 40.9  & 1.5    & 18.5      & 20.0       & 36.9     & 1.0       & 22.5         & 28.5        \\
                            & falcon-40b-instruct   & 76.5  & 28.1   & 18.9      & 24.0       & 54.6     & 10.6      & 23.4         & 24.4        \\
                            & GPT-3                 & 93.2  & 28.9   & 25.6      & 18.0       & 72.7     & 24.4      & 27.7         & 23.2        \\ 
                            & GPT-3~\cite{ribeiro2022street}\textsuperscript{*} & 72.9 & 34.8 & 40.2 & 19.0 & - & - & - & - \\ \midrule
\multirow{4}{*}{\rotatebox{90}{\Sim\ code}}       & BLOOMZ                & 61.5  & 0.7    & 23.6      & 26.0       & 52.6     & 1.5       & 23.3         & 24.0        \\
                            & BLOOMZ -TCC (ours)    & 76.2  & 1.9    & 21.7      & 28.0       & 61.1     & 1.3       & 25.4         & 24.8        \\
                            & falcon-40b-instruct   & 81.5  & 19.6   & 24.4      & 34.0       & 43.9     & 4.6       & 23.5         & 26.8        \\
                            & GPT-3                 & 99.1  & 45.2   & 37.8      & 32.0       & 94.2     & 26.9      & 32.8         & 24.8        \\ \bottomrule
\end{tabular}
\caption{Full results (STREET and xSTREET average) for all models, tasks, prompt formats, and languages. Rows are grouped by prompt type and model, while columns are grouped by the subtask and the language. \\ \small{\textsuperscript{*}We include the reported results of \citet{ribeiro2022street} for STREET. This is for reference, as we cannot reproduce their exact prompts and examples chosen (and so cannot run on xSTREET).} }
\label{tab:full results}
\end{table*}

\begin{table*}[ht!]
\small
\centering
\setlength{\tabcolsep}{1.5pt}
\begin{tabular}{@{}ll|rrrrr|rrrrr|rrrrr|rrrrr@{}}
 & \multicolumn{1}{r|}{} & \multicolumn{5}{c|}{ARC} & \multicolumn{5}{c|}{GSM8K} & \multicolumn{5}{c|}{AQUA\_RAT} & \multicolumn{5}{c}{AR\_LSAT} \\
 & Model & es & zh & ar & ja & ru & es & zh & ar & ja & ru & es & zh & ar & ja & ru & es & zh & ar & ja & ru \\ \midrule
\multirow{4}{*}{\rotatebox{90}{direct}} & BLOOMZ & 46.5 & 41.2 & 57.4 & 33.5 & 30.0 & 2.2 & 1.9 & 1.1 & 1.1 & 2.2 & 19.7 & 24.8 & 21.3 & 24.0 & 15.7 & 22.0 & 24.0 & 16.0 & 24.0 & 18.0 \\
 & BLOOMZ -TCC & 63.2 & 44.7 & 68.5 & 31.2 & 37.6 & 1.9 & 1.5 & 2.2 & 1.5 & 1.9 & 22.0 & 25.6 & 19.7 & 22.0 & 20.1 & 18.0 & 28.0 & 26.0 & 24.0 & 20.0 \\
 & falcon-40b-instruct & 34.7 & 32.4 & 30.3 & 31.8 & 30.9 & 10.0 & 6.7 &  3.3 &  4.8 &  5.6 & 24.8 & 24.0 & 25.2 & 24.8 & 24.0 & 26.0 & 22.0 & 24.0 & 26.0 & 22.0  \\
 & GPT-3 & 97.4 & 95.0 & 81.8 & 90.6 & 88.5 & 3.0 & 1.5 & 0.0 & 1.5 & 1.9 & 25.2 & 24.8 & 25.6 & 26.0 & 26.4 & 24.0 & 24.0 & 24.0 & 24.0 & 26.0 \\ \midrule
\multirow{4}{*}{\rotatebox{90}{linearized}} & BLOOMZ & 35.0 & 37.1 & 35.3 & 29.1 & 30.9 & 2.2 & 1.1 & 0.4 & 0.4 & 0.4 & 22.0 & 26.8 & 23.6 & 24.0 & 25.6 & 24.0 & 26.0 & 24.0 & 20.0 & 26.0 \\
 & BLOOMZ-TCC & 44.7 & 40.3 & 39.7 & 30.9 & 29.1 & 1.5 & 0.4 & 1.5 & 0.4 & 1.1 & 17.3 & 21.7 & 21.3 & 22.8 & 24.0 & 14.0 & 30.0 & 28.0 & 28.0 & 28.0 \\
 & falcon-40b-instruct & 77.4 & 65.6 & 34.7 & 49.4 & 45.9 &  23.3 & 16.3 & 2.2 &  7.0 &  4.4 &  25.2 & 25.2 & 21.7 & 22.0 & 22.8 &  18.0 & 26.0 & 24.0 & 18.0 & 36.0\\
 & GPT-3 & 83.5 & 75.6 & 57.4 & 69.7 & 77.4 & 26.3 & 29.6 & 19.3 & 17.0 & 30.0 & 27.6 & 27.6 & 29.5 & 26.4 & 27.6 & 26.0 & 28.0 & 20.0 & 26.0 & 16.0 \\ \midrule
\multirow{4}{*}{\rotatebox{90}{\Sim\ code}} & BLOOMZ & 57.9 & 67.6 & 58.8 & 45.6 & 32.9 & 0.7 & 1.5 & 2.2 & 2.2 & 0.7 & 21.3 & 25.6 & 23.6 & 22.4 & 23.6 & 26.0 & 30.0 & 30.0 & 12.0 & 22.0 \\
 & BLOOMZ-TCC & 70.3 & 74.4 & 65.0 & 54.7 & 40.9 & 0.4 & 1.5 & 1.5 & 2.2 & 0.7 & 27.2 & 24.0 & 28.3 & 22.8 & 24.8 & 26.0 & 34.0 & 16.0 & 30.0 & 18.0 \\
 & falcon-40b-instruct & 57.1 & 55.3 & 30.3 & 44.1 & 32.6 &  8.1 &  8.1 &  1.9 & 2.6  &  2.2 &  26.0 & 24.4 & 23.2 & 20.5& 23.2 &  32.0 & 26.0 & 26.0 & 24.0 & 26.0 \\
 & GPT-3 & 96.5 & 96.5 & 90.3 & 94.1 & 93.5 & 37.4 & 28.1 & 18.5 & 21.9 & 28.5 & 36.6 & 32.7 & 31.1 & 32.7 & 31.1 & 26.0 & 28.0 & 26.0 & 18.0 & 26.0 \\ \bottomrule
\end{tabular}
\caption{Per-language xSTREET results for all models, tasks, prompt formats, and languages. Rows are grouped by prompt type and model, while columns are grouped by the subtask and the language.}
\label{tab:per_lang_results}
\end{table*}

\section{Additional Experiments}
\label{sec:ablations}
We perform 2 additional ablation experiments below.

\subsection{Finetuning on \Sim\ Code Prompts}
\label{sec:direct_ft}
We experiment with directly finetuning on \Sim\ code prompts (all 6 languages), so as to have a model that can perform the \Sim-formatted reasoning without in-context examples. We use the same hyperparameter and configuration from \S\ref{sec:ft}, again using LoRA to fine-tune a subset of the 7B model's parameters, but omitting the data replay to maximize performance. We train one model for all tasks, and all 6 languages. This differentiates our \Sim\ finetuned model from the linearized finetuned model of~\citet{ribeiro2022street}, which finetunes a separate T5 (0.8B) model for each task with full finetuning. 

As multilingual trends remain similar, we will just discuss English results (STREET).%
Using BLOOMZ-\tcc\ or BLOOMZ as the base model does not make a difference. The \Sim\ finetuned models achieves 85.9 (vs. 76.2) on ARC; the other 3 tasks are still near random chance. We suspect that performing full finetuning instead of LoRA should overcome this, and omit this experiment due to resource constraints.

\subsection{Improving the Code Comment Quality of \tcc}
\label{sec:tcc_filtering}
Our initial and used version of \tcc, as described in the main text, simply took 30k source code files from The Stack, then filtered down to 20k files, using criteria of >5 comments, of which >50\% of comments in English. 

To validate the quality of the translated code files, we recruited human annotators who were proficient  programmers, and native speakers of each language. While overall, translations were judged to be reasonable, the main feedback points were:
\begin{itemize}
    \item Some files had non-useful comments, such as long copyright statements in the header, or linting messages.
    \item Some files had comments which were actually commented-out code (i.e. unused functions).
    \item Terms related to programming, or referencing function or variable names in the code, were often mistranslated, if they should have been translated at all.
\end{itemize}

We leave the last point to future work, as we used an online MT API, and programming-specific MT is out of scope. For the other two, we tried to develop a version of \tcc\ to specifically select files which have plenty of meaningful comments. We describe this filtering experiment below.

We now consider the entire Stack dataset,\footnote{\url{https://huggingface.co/datasets/bigcode/the-stack-dedup}}, instead of just 30k from the official small subset. As the Stack totals 6 TB, we considered only the first 3 million examples (1m each for Java, Python, JavaScript). Our scripts performed the following steps in order:

\begin{enumerate}
    \item Delete copyrights, headers, linting comments from files.
    \item Keep only those files with >1 standard deviation of number of comments:number of lines ratio. Note that 1 comment can span multiple lines.
    \item Keep only those files with >5 comments.
\end{enumerate}

This resulted in about 250K examples. We performed the code comment extraction and translation process, and termed the resulting dataset \tcc-v2. We then applied the finetuning recipe as we did with the original \tcc; furthermore, we keep the data size consistent as the original, using a 67k subset of \tcc-v2. The resulting BLOOMZ-\tcc-v2 models had similar downstream reasoning performance as BLOOMZ-\tcc, and therefore we did not use it in the main text.

We hypothesize this experiment did not improve performance because, the program code plays a much bigger role in LLM's reasoning abilities than the comments.

%% file: gsm8k_en_ex.tex
\begin{table*}[ht!]
\centering
\tiny
\setlength{\tabcolsep}{4pt}

\begin{tabularx}{\linewidth}{lX} \toprule
\Sim\ code input & facts = {[}'Raymond and Samantha are cousins.', 'Raymond was born 6 years before Samantha.', 'Raymond had a son at the age of 23.', 'If Samantha is now 31,'{]}\newline question = "how many years ago was Raymond's son born?"\newline answer = None\newline \# call select\_facts and infer\_new\_fact until answer is set\newline selected = select\_facts(facts) \# \\ \midrule
Linearized input & QUESTION = sent1: Raymond and Samantha are cousins. sent2: Raymond was born 6 years before Samantha. sent3: Raymond had a son at the age of 23. sent4: If Samantha is now 31, sent5: how many years ago was Raymond's son born? \newline PROOF = \\ \bottomrule
\end{tabularx}

\begin{tabularx}{\textwidth}{lXXXX} \toprule
& Gold & GPT-3 & BLOOMZ-tcc & BLOOMZ \\ \midrule
    \rotatebox{90}{\hspace{-10em}\Sim\ code output} 
    & {[}facts{[}1{]}, facts{[}2{]}{]}\newline new\_fact = infer\_new\_fact(selected) \# "When Raymond's son was born Samantha was 23 - 6 = 17 years old."\newline is\_solved(new\_fact, question) \# False\newline facts.append(new\_fact)\newline \newline selected = select\_facts(facts) \# {[}facts{[}3{]}, facts{[}3{]}, facts{[}4{]}{]}\newline new\_fact = infer\_new\_fact(selected) \# "Thus it has been 31 - 17 = 14 years since Raymond's son was born."\newline is\_solved(new\_fact, question) \# False\newline facts.append(new\_fact)\newline \newline selected = select\_facts(facts) \# {[}facts{[}3{]}, facts{[}5{]}{]}\newline new\_fact = infer\_new\_fact(selected) \# 'The answer is 14'\newline is\_solved(new\_fact, question) \# True\newline answer = extract\_answer(new\_fact) \# \textcolor{OliveGreen}{'14'}
    
    & {[}facts{[}1{]}, facts{[}2{]}{]}\newline new\_fact = infer\_new\_fact(selected) \# 'Raymond's son was born 23 - 6 = 17 years ago.'\newline is\_solved(new\_fact, question) \# False\newline facts.append(new\_fact)\newline \newline selected = select\_facts(facts) \# {[}facts{[}3{]}, facts{[}4{]}{]}\newline new\_fact = infer\_new\_fact(selected) \# 'Samantha is 31, so Raymond's son was born 31 - 17 = 14 years ago.'\newline is\_solved(new\_fact, question) \# False\newline facts.append(new\_fact)\newline \newline selected = select\_facts(facts) \# {[}facts{[}3{]}, facts{[}5{]}{]}\newline new\_fact = infer\_new\_fact(selected) \# 'The answer is 14'\newline is\_solved(new\_fact, question) \# True\newline answer = extract\_answer(new\_fact) \# \textcolor{OliveGreen}{'14'}
    
    & {[}facts{[}0{]}{]}\newline new\_fact = infer\_new\_fact(selected) \# facts{[}2{]}\newline is\_solved(new\_fact, question) \# False\newline facts.append(new\_fact)\newline \newline selected = select\_facts(facts) \# {[}facts{[}0{]}, facts{[}2{]}{]}\newline new\_fact = infer\_new\_fact(selected) \# facts{[}2{]}\newline is\_solved(new\_fact, question) \# True\newline facts.append(new\_fact)\newline \newline selected = select\_facts(facts) \# {[}question, facts{[}2{]}{]}\newline new\_fact = infer\_new\_fact(selected) \# facts{[}2{]}\newline is\_solved(new\_fact, question) \# True\newline answer = extract\_answer(new\_fact) \# \textcolor{BrickRed}{'23'}
    
    & new\_fact = infer\_new\_fact(selected) \# facts{[}2{]}\newline is\_solved(new\_fact, question) \# False\newline facts.append(new\_fact)\newline \newline \textcolor{gray}{\textit{\#\#\# \newline}} selected = select\_facts(facts) \#   {[}facts{[}0{]}, facts{[}1{]}, facts{[}2{]}, facts{[}3{]}{]}\newline new\_fact = infer\_new\_fact(selected) \# facts{[}2{]}\newline is\_solved(new\_fact, question) \# True\newline facts.append(new\_fact) \textcolor{gray}{\textit{\newline \#\#\# repeated 4 times, without a parseable answer}} \\ \midrule
    
    \rotatebox[origin=c]{90}{\hspace{-10em}linearized output} & sent2 \& sent3 -\textgreater int1: When Raymond's son was born Samantha was 23 - 6 = 17 years old.; int1 \& sent4 \& sent5 -\textgreater int2: Thus it has been 31 - 17 = 14 years since Raymond's son was born.; int2 \& sent5 -\textgreater int3: The answer is \textcolor{OliveGreen}{14};
    
    & sent2 \& sent4 -\textgreater int1: Raymond was born 6 years before Samantha, so Samantha is now 31 - 6 = 25 years old.; int1 \& sent3 -\textgreater int2: Raymond had a son at the age of 23, so Raymond's son was born 23 - 6 = 17 years ago.; int2 \& sent5 -\textgreater int3: The answer is \textcolor{BrickRed}{17};
    
    & sent1 \& sent2 -\textgreater int1: Raymond was born 6 years before Samantha, so Raymond was born 6 years before Samantha.; int1 \& sent2 \& sent3 -\textgreater int2: Raymond had a son at the age of 23, so Raymond was 23 years old when he had a son.; int1 \& int2 \& sent3 -\textgreater int3: Samantha is now 31, so Samantha was 31 years old when she had a son.; int1 \& int3 \& sent4 -\textgreater int5: Thus, Raymond had a son 6 years ago.; int1 \& int4 \& sent5 -\textgreater int6: The answer is \textcolor{BrickRed}{6} years.;
    
    & sent1 \& sent2 \& sent3 -\textgreater int1: Raymond was born 6 years before Samantha, so Raymond was born 6 years before Samantha's son was born.; int1 \& int2 \& sent3 \& sent4 -\textgreater int2: Samantha is now 31, so Samantha's son was born 6 years ago.; int1 \& int2 \& int3 \& sent4 -\textgreater int5: The answer is \textcolor{BrickRed}{6} years.; \\ \bottomrule
\end{tabularx}
\caption{Case study for a GSM8K entry, in English. The correct answer is `14' (in green). Only GPT-3 with \Sim\ prompts gets it correct.}
\label{tab:gsm8k_en_study}
\end{table*}

%% file: arc_arabic.tex
\begin{table*}[ht!]
\centering
\tiny
\setlength{\tabcolsep}{4pt}
\includegraphics[]{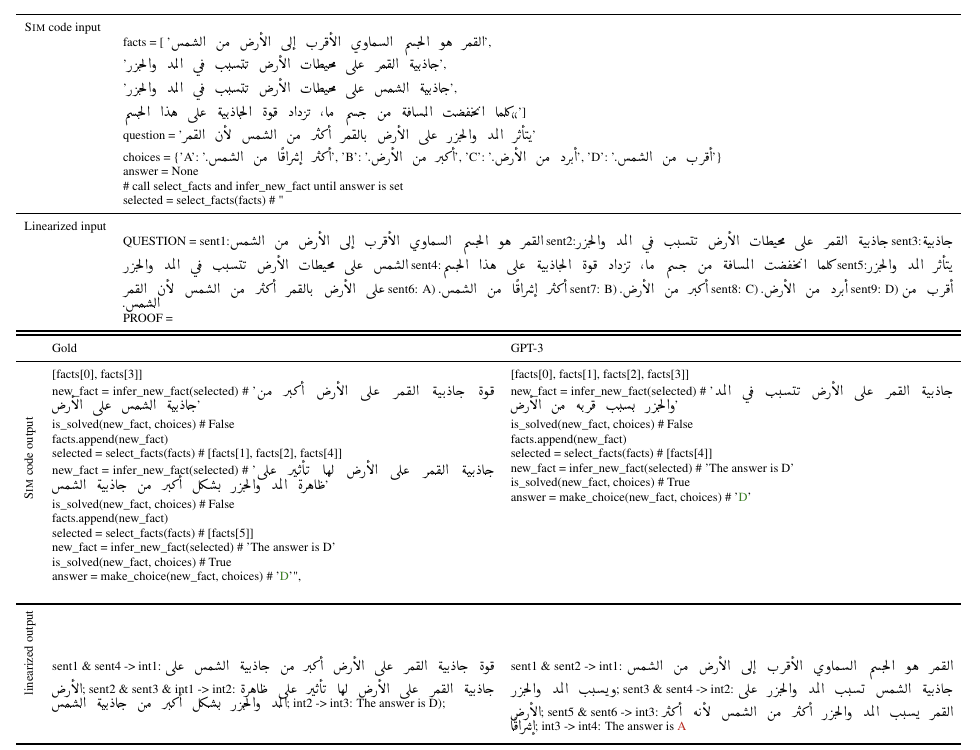}
\caption{Case study for a ARC entry, in Arabic. The correct answer is `D' (in green). Only GPT-3 with \Sim\ prompts gets it correct.}
\label{tab:arc_ar_study}
\end{table*}

%% file: acl_latex.bbl
\begin{thebibliography}{24}
\expandafter\ifx\csname natexlab\endcsname\relax\def\natexlab#1{#1}\fi

\bibitem[{Ahuja et~al.(2023)Ahuja, Hada, Ochieng, Jain, Diddee, Maina, Ganu, Segal, Axmed, Bali et~al.}]{ahuja2023mega}
Kabir Ahuja, Rishav Hada, Millicent Ochieng, Prachi Jain, Harshita Diddee, Samuel Maina, Tanuja Ganu, Sameer Segal, Maxamed Axmed, Kalika Bali, et~al. 2023.
\newblock Mega: Multilingual evaluation of generative ai.
\newblock \emph{arXiv preprint arXiv:2303.12528}.

\bibitem[{Almazrouei et~al.(2023)Almazrouei, Alobeidli, Alshamsi, Cappelli, Cojocaru, Debbah, Étienne Goffinet, Hesslow, Launay, Malartic, Mazzotta, Noune, Pannier, and Penedo}]{almazrouei2023falcon}
Ebtesam Almazrouei, Hamza Alobeidli, Abdulaziz Alshamsi, Alessandro Cappelli, Ruxandra Cojocaru, Mérouane Debbah, Étienne Goffinet, Daniel Hesslow, Julien Launay, Quentin Malartic, Daniele Mazzotta, Badreddine Noune, Baptiste Pannier, and Guilherme Penedo. 2023.
\newblock \href {http://arxiv.org/abs/2311.16867} {The falcon series of open language models}.

\bibitem[{Bi et~al.(2023)Bi, Chen, Jiang, Xiong, Guo, Chen, and Zhang}]{bi2023codekgc}
Zhen Bi, Jing Chen, Yinuo Jiang, Feiyu Xiong, Wei Guo, Huajun Chen, and Ningyu Zhang. 2023.
\newblock Codekgc: Code language model for generative knowledge graph construction.
\newblock \emph{arXiv preprint arXiv:2304.09048}.

\bibitem[{Chen et~al.(2023)Chen, Ma, Wang, and Cohen}]{chen2023program}
Wenhu Chen, Xueguang Ma, Xinyi Wang, and William~W Cohen. 2023.
\newblock Program of thoughts prompting: Disentangling computation from reasoning for numerical reasoning tasks.
\newblock \emph{Transactions on Machine Learning Research}.

\bibitem[{Cobbe et~al.(2021)Cobbe, Kosaraju, Bavarian, Chen, Jun, Kaiser, Plappert, Tworek, Hilton, Nakano et~al.}]{cobbe2021training}
Karl Cobbe, Vineet Kosaraju, Mohammad Bavarian, Mark Chen, Heewoo Jun, Lukasz Kaiser, Matthias Plappert, Jerry Tworek, Jacob Hilton, Reiichiro Nakano, et~al. 2021.
\newblock Training verifiers to solve math word problems.
\newblock \emph{arXiv preprint arXiv:2110.14168}.

\bibitem[{Creswell et~al.(2023)Creswell, Shanahan, and Higgins}]{creswellSelectionInferenceExploitingLarge2023}
Antonia Creswell, Murray Shanahan, and Irina Higgins. 2023.
\newblock \href {https://openreview.net/forum?id=3Pf3Wg6o-A4} {Selection-{{Inference}}: {{Exploiting Large Language Models}} for {{Interpretable Logical Reasoning}}}.
\newblock In \emph{The {{Eleventh International Conference}} on {{Learning Representations}}}.

\bibitem[{Dong et~al.(2023)Dong, Martin, and Callison-Burch}]{dong2023corrpus}
Yijiang Dong, Lara Martin, and Chris Callison-Burch. 2023.
\newblock Corrpus: Code-based structured prompting for neurosymbolic story understanding.
\newblock In \emph{Findings of the Association for Computational Linguistics: ACL 2023}, pages 13152--13168.

\bibitem[{Hendy et~al.(2023)Hendy, Abdelrehim, Sharaf, Raunak, Gabr, Matsushita, Kim, Afify, and Awadalla}]{Hendy2023HowGA}
Amr Hendy, Mohamed~Gomaa Abdelrehim, Amr Sharaf, Vikas Raunak, Mohamed Gabr, Hitokazu Matsushita, Young~Jin Kim, Mohamed Afify, and Hany~Hassan Awadalla. 2023.
\newblock \href {https://api.semanticscholar.org/CorpusID:257038384} {How good are gpt models at machine translation? a comprehensive evaluation}.
\newblock \emph{ArXiv}, abs/2302.09210.

\bibitem[{Hu et~al.(2021)Hu, Wallis, Allen-Zhu, Li, Wang, Wang, Chen et~al.}]{hu2021lora}
Edward~J Hu, Phillip Wallis, Zeyuan Allen-Zhu, Yuanzhi Li, Shean Wang, Lu~Wang, Weizhu Chen, et~al. 2021.
\newblock Lora: Low-rank adaptation of large language models.
\newblock In \emph{International Conference on Learning Representations}.

\bibitem[{Kocetkov et~al.(2022)Kocetkov, Li, Allal, Li, Mou, Ferrandis, Jernite, Mitchell, Hughes, Wolf et~al.}]{kocetkov2022stack}
Denis Kocetkov, Raymond Li, Loubna~Ben Allal, Jia Li, Chenghao Mou, Carlos~Mu{\~n}oz Ferrandis, Yacine Jernite, Margaret Mitchell, Sean Hughes, Thomas Wolf, et~al. 2022.
\newblock The stack: 3 tb of permissively licensed source code.
\newblock \emph{arXiv preprint arXiv:2211.15533}.

\bibitem[{Li et~al.(2024)Li, Haider, and Callison-Burch}]{li2024land}
Bryan Li, Samar Haider, and Chris Callison-Burch. 2024.
\newblock \href {http://arxiv.org/abs/2305.14610} {This land is {Your, My} land: Evaluating geopolitical biases in language models}.

\bibitem[{Liang et~al.(2023)Liang, Bommasani, Lee, Tsipras, Soylu, Yasunaga, Zhang, Narayanan, Wu, Kumar, Newman, Yuan, Yan, Zhang, Cosgrove, Manning, R'e, Acosta-Navas, Hudson, Zelikman, Durmus, Ladhak, Rong, Ren, Yao, Wang, Santhanam, Orr, Zheng, Yuksekgonul, Suzgun, Kim, Guha, Chatterji, Khattab, Henderson, Huang, Chi, Xie, Santurkar, Ganguli, Hashimoto, Icard, Zhang, Chaudhary, Wang, Li, Mai, Zhang, and Koreeda}]{Liang2023HolisticEO}
Percy Liang, Rishi Bommasani, Tony Lee, Dimitris Tsipras, Dilara Soylu, Michihiro Yasunaga, Yian Zhang, Deepak Narayanan, Yuhuai Wu, Ananya Kumar, Benjamin Newman, Binhang Yuan, Bobby Yan, Ce~Zhang, Christian Cosgrove, Christopher~D. Manning, Christopher R'e, Diana Acosta-Navas, Drew~A. Hudson, E.~Zelikman, Esin Durmus, Faisal Ladhak, Frieda Rong, Hongyu Ren, Huaxiu Yao, Jue Wang, Keshav Santhanam, Laurel~J. Orr, Lucia Zheng, Mert Yuksekgonul, Mirac Suzgun, Nathan~S. Kim, Neel Guha, Niladri~S. Chatterji, Omar Khattab, Peter Henderson, Qian Huang, Ryan Chi, Sang~Michael Xie, Shibani Santurkar, Surya Ganguli, Tatsunori Hashimoto, Thomas~F. Icard, Tianyi Zhang, Vishrav Chaudhary, William Wang, Xuechen Li, Yifan Mai, Yuhui Zhang, and Yuta Koreeda. 2023.
\newblock \href {https://api.semanticscholar.org/CorpusID:253553585} {Holistic evaluation of language models}.
\newblock \emph{Annals of the New York Academy of Sciences}, 1525:140 -- 146.

\bibitem[{Madaan et~al.(2022)Madaan, Zhou, Alon, Yang, and Neubig}]{madaan-etal-2022-language}
Aman Madaan, Shuyan Zhou, Uri Alon, Yiming Yang, and Graham Neubig. 2022.
\newblock \href {https://doi.org/10.18653/v1/2022.emnlp-main.90} {Language models of code are few-shot commonsense learners}.
\newblock In \emph{Proceedings of the 2022 Conference on Empirical Methods in Natural Language Processing}, pages 1384--1403, Abu Dhabi, United Arab Emirates. Association for Computational Linguistics.

\bibitem[{Muennighoff et~al.(2022)Muennighoff, Wang, Sutawika, Roberts, Biderman, Scao, Bari, Shen, Yong, Schoelkopf et~al.}]{muennighoff2022crosslingual}
Niklas Muennighoff, Thomas Wang, Lintang Sutawika, Adam Roberts, Stella Biderman, Teven~Le Scao, M~Saiful Bari, Sheng Shen, Zheng-Xin Yong, Hailey Schoelkopf, et~al. 2022.
\newblock Crosslingual generalization through multitask finetuning.
\newblock \emph{arXiv preprint arXiv:2211.01786}.

\bibitem[{Ni et~al.(2022)Ni, Inala, Wang, Polozov, Meek, Radev, and Gao}]{Ni2022LearningMR}
Ansong Ni, Jeevana~Priya Inala, Chenglong Wang, Oleksandr Polozov, Christopher Meek, Dragomir~R. Radev, and Jianfeng Gao. 2022.
\newblock \href {https://api.semanticscholar.org/CorpusID:257019561} {Learning math reasoning from self-sampled correct and partially-correct solutions}.
\newblock In \emph{International Conference on Learning Representations}.

\bibitem[{Patel et~al.(2022)Patel, Li, Rasooli, Constant, Raffel, and Callison-Burch}]{patel2022bidirectional}
Ajay Patel, Bryan Li, Mohammad~Sadegh Rasooli, Noah Constant, Colin Raffel, and Chris Callison-Burch. 2022.
\newblock Bidirectional language models are also few-shot learners.
\newblock In \emph{The Eleventh International Conference on Learning Representations}.

\bibitem[{Ribeiro et~al.(2022)Ribeiro, Wang, Ma, Zhu, Dong, Kong, Burger, Ramos, Wang, Karypis et~al.}]{ribeiro2022street}
Danilo~Neves Ribeiro, Shen Wang, Xiaofei Ma, Henghui Zhu, Rui Dong, Deguang Kong, Juliette Burger, Anjelica Ramos, William~Yang Wang, George Karypis, et~al. 2022.
\newblock Street: A multi-task structured reasoning and explanation benchmark.
\newblock In \emph{The Eleventh International Conference on Learning Representations}.

\bibitem[{Shi et~al.(2022)Shi, Suzgun, Freitag, Wang, Srivats, Vosoughi, Chung, Tay, Ruder, Zhou et~al.}]{shi2022language}
Freda Shi, Mirac Suzgun, Markus Freitag, Xuezhi Wang, Suraj Srivats, Soroush Vosoughi, Hyung~Won Chung, Yi~Tay, Sebastian Ruder, Denny Zhou, et~al. 2022.
\newblock Language models are multilingual chain-of-thought reasoners.
\newblock In \emph{The Eleventh International Conference on Learning Representations}.

\bibitem[{Suzgun et~al.(2022)Suzgun, Scales, Scharli, Gehrmann, Tay, Chung, Chowdhery, Le, hsin Chi, Zhou, and Wei}]{Suzgun2022ChallengingBT}
Mirac Suzgun, Nathan Scales, Nathanael Scharli, Sebastian Gehrmann, Yi~Tay, Hyung~Won Chung, Aakanksha Chowdhery, Quoc~V. Le, Ed~Huai hsin Chi, Denny Zhou, and Jason Wei. 2022.
\newblock \href {https://api.semanticscholar.org/CorpusID:252917648} {Challenging big-bench tasks and whether chain-of-thought can solve them}.
\newblock In \emph{Annual Meeting of the Association for Computational Linguistics}.

\bibitem[{Wei et~al.(2022{\natexlab{a}})Wei, Tay, Bommasani, Raffel, Zoph, Borgeaud, Yogatama, Bosma, Zhou, Metzler et~al.}]{wei2022emergent}
Jason Wei, Yi~Tay, Rishi Bommasani, Colin Raffel, Barret Zoph, Sebastian Borgeaud, Dani Yogatama, Maarten Bosma, Denny Zhou, Donald Metzler, et~al. 2022{\natexlab{a}}.
\newblock Emergent abilities of large language models.
\newblock \emph{Transactions on Machine Learning Research}.

\bibitem[{Wei et~al.(2022{\natexlab{b}})Wei, Wang, Schuurmans, Bosma, Xia, Chi, Le, Zhou et~al.}]{wei2022chain}
Jason Wei, Xuezhi Wang, Dale Schuurmans, Maarten Bosma, Fei Xia, Ed~Chi, Quoc~V Le, Denny Zhou, et~al. 2022{\natexlab{b}}.
\newblock Chain-of-thought prompting elicits reasoning in large language models.
\newblock \emph{Advances in Neural Information Processing Systems}, 35:24824--24837.

\bibitem[{Wolf et~al.(2020)Wolf, Debut, Sanh, Chaumond, Delangue, Moi, Cistac, Rault, Louf, Funtowicz, Davison, Shleifer, von Platen, Ma, Jernite, Plu, Xu, Scao, Gugger, Drame, Lhoest, and Rush}]{wolf-etal-2020-transformers}
Thomas Wolf, Lysandre Debut, Victor Sanh, Julien Chaumond, Clement Delangue, Anthony Moi, Pierric Cistac, Tim Rault, Rémi Louf, Morgan Funtowicz, Joe Davison, Sam Shleifer, Patrick von Platen, Clara Ma, Yacine Jernite, Julien Plu, Canwen Xu, Teven~Le Scao, Sylvain Gugger, Mariama Drame, Quentin Lhoest, and Alexander~M. Rush. 2020.
\newblock \href {https://www.aclweb.org/anthology/2020.emnlp-demos.6} {Transformers: State-of-the-art natural language processing}.
\newblock In \emph{Proceedings of the 2020 Conference on Empirical Methods in Natural Language Processing: System Demonstrations}, pages 38--45, Online. Association for Computational Linguistics.

\bibitem[{Yao et~al.(2023)Yao, Aminabadi, Ruwase, Rajbhandari, Wu, Awan, Rasley, Zhang, Li, Holmes, Zhou, Wyatt, Smith, Kurilenko, Qin, Tanaka, Che, Song, and He}]{yao2023dschat}
Zhewei Yao, Reza~Yazdani Aminabadi, Olatunji Ruwase, Samyam Rajbhandari, Xiaoxia Wu, Ammar~Ahmad Awan, Jeff Rasley, Minjia Zhang, Conglong Li, Connor Holmes, Zhongzhu Zhou, Michael Wyatt, Molly Smith, Lev Kurilenko, Heyang Qin, Masahiro Tanaka, Shuai Che, Shuaiwen~Leon Song, and Yuxiong He. 2023.
\newblock {DeepSpeed-Chat: Easy, Fast and Affordable RLHF Training of ChatGPT-like Models at All Scales}.
\newblock \emph{arXiv preprint arXiv:2308.01320}.

\bibitem[{Zhang et~al.(2023)Zhang, Xu, Yang, Zhou, You, Arora, and Callison-Burch}]{zhang-etal-2023-causal}
Li~Zhang, Hainiu Xu, Yue Yang, Shuyan Zhou, Weiqiu You, Manni Arora, and Chris Callison-Burch. 2023.
\newblock \href {https://doi.org/10.18653/v1/2023.findings-eacl.31} {Causal reasoning of entities and events in procedural texts}.
\newblock In \emph{Findings of the Association for Computational Linguistics: EACL 2023}, pages 415--431, Dubrovnik, Croatia. Association for Computational Linguistics.

\end{thebibliography}
